\documentclass[conference]{IEEEtran}
\ifCLASSINFOpdf
 \usepackage[pdftex]{graphicx}
\else
\fi
\ifCLASSOPTIONcompsoc
  \usepackage[caption=false,font=normalsize,labelfont=sf,textfont=sf]{subfig}
\else
  \usepackage[caption=false,font=footnotesize]{subfig}
\fi
\usepackage[utf8]{inputenc} 
\usepackage[T1]{fontenc}    
\usepackage{url}            
\usepackage{booktabs}       
\usepackage{amsfonts}       
\usepackage{nicefrac}       
\usepackage{microtype}      
\usepackage{cite}

\usepackage{amsmath, amssymb, amsthm, tikz, pgfplots}

\newcommand*{\Cdot}{\raisebox{-0.4ex}{\scalebox{1.15}{$\cdot$}}}

\begin{document}
%
\title{HNP3: A Hierarchical Nonparametric Point Process for Modeling Content Diffusion over Social Media}


\author{\IEEEauthorblockN{Seyed Abbas Hosseini,
Ali Khodadadi,
Soheil Arabzade  and
Hamid R. Rabiee}
\IEEEauthorblockA{AICT Innovation Center, Department of Computer Engineering\\
Sharif University of Technology, Tehran, Iran\\
Email: \{a\_hosseini, khodadadi, arabzade\}@ce.sharif.edu, rabiee@sharif.edu}}


%


\maketitle

\begin{abstract}
This paper introduces a novel framework for modeling temporal events with complex longitudinal dependency that are generated by dependent sources. 
This framework takes advantage of multidimensional point processes for modeling time of events. The intensity function of the proposed process is a mixture of intensities, and its complexity grows with the complexity of temporal patterns of data. Moreover, it utilizes a hierarchical dependent nonparametric approach to model marks of events. These capabilities allow the proposed model to adapt its temporal and topical complexity according to the complexity of data, which makes it a suitable candidate for real world scenarios. An online inference algorithm is also proposed that makes the framework applicable to a vast range of applications. The framework is applied to a real world application, modeling the diffusion of contents over networks. Extensive experiments reveal the effectiveness of the proposed framework in comparison with state-of-the-art methods.
\end{abstract}


%
\IEEEpeerreviewmaketitle

\section{Introduction}
A huge amount of information in the form of news, photos, and tweets propagates through social media and networks. Analyzing these information, can help us understand the users' interests and their influence on each other. This kind of knowledge help us to understand how applications such as online advertising operate through incentivizing users \cite{farajtabar2014}. Considering the temporal dynamics of the different topics discussed over the networks can immensely help marketers run more effective campaigns \cite{UncoverTopicSensitive}. Therefore, there has been a large amount of research on the analysis and modeling of the content being shared over social networks to extract users preferences and the amount of their influence on each other.
  
  
Users of social networks often share information in one form or another. The content and temporal characteristics of what is shared as well as the relations among members are the three main sources that allow us to identify users' interests over time and their influence characteristics. 
  
Modeling the content that is shared on a network over time has many challenges. This content covers a wide range of topics. Each of these topics emerge at some point, become popular to some extent,  influence some parts of the network, and eventually fade out. However, topics propagate with different rates. Therefore, we need a flexible model that can not only represent the dynamics of topic popularity, but also model the diversity and diffusion rate of topics over time. 
  
There exist many dependent nonparametric models for the diversity and dynamics of the topics in a text stream. Two nonparametric topic-cluster models were introduced in \cite{AhmedOnlineInference,ahmed2012timeline} that cluster news based on their topics and infer the number of clusters, concurrently. These models have two main drawbacks. First, they only model a single source of information and hence do not consider the impact of different sources on each other. Second, they only consider time as a covariate, while modeling the time of news events can enhance accuracy of the method in finding topics and also the influence of users on each other. The authors in \cite{DirichletHawkes} have recently proposed a nonparametric point process that jointly models both topic and time of the events. However, this method assumes that the data are generated by a single source and hence is not applicable of analyzing events over a network.
  
A rich literature exists on modeling information diffusion over networks \cite{Iwata2013,Zhou2013a,tran2015netcodec}. These methods model the time of events using a point process such as Hawkes process \cite{liniger2009multivariate} but fall short of considering the content. Some recent methods such as \cite{ViralDiffusion,UncoverTopicSensitive} consider the content of the event but assume that the topics are already known. The authors in \cite{HawkesTopic} have recently proposed a method that jointly models the content and the time of events to infer the topic of the events and the influence network. However, this method assumes that the number of topics is bounded and known, and also the time and topics of events are assumed to be independent. These assumptions are not valid in social and information networks, where new topics arise over time, and the rate of diffusion of different content is heavily dependent on their topic \cite{du2012learning}.

In this paper, we propose a nonparametric point process that jointly models the topic and time of events generated by the users of a network, infers the users' influence on each other, and their dynamic interests over time in an online manner. In this model, each topic has a specific temporal dynamic which determines its diffusion rate through the network. The model is nonparametric and adapts the number of topics according to the complexity of data. In summary, we make the following contributions:
\begin{itemize}
	\item We introduce a nonparametric multidimensional point process that can jointly model the time and topic of events for a set of dependent sources. This model permits topics to be shared among different sources using a hierarchical structure and is able to adapt its complexity according to the complexity of data.
	\item Our model provides a dynamic hierarchical clustering over the events, in three levels. In the first level, the events are clustered based on the root event that has triggered them. In the second level, for each user, the root events of each cluster are grouped based on their topic and temporal dynamics. Finally, in the third level, the topics of events are clustered irrespective of their user. This clustering allows us to better understand the interests of users and also the trending of topics over the network.
	\item We propose an efficient online inference algorithm based on the collapsed Sequential Monte Carlo that relies on marginalizing global latent variables to speed up the inference process. The inference algorithm is online, which makes it a suitable choice for real applications with millions of events.
	\item We conduct several experiments on synthetic and real world datasets to evaluate the performance of our model. To this end, we collected a dataset consisting of 100,000 news articles published over 3 months by 100 news websites.
\end{itemize}

The remainder of this paper is organized as follows. In section \ref{sec:SettingsAndBackground} we briefly review the necessary background. Details of the proposed method is discussed in section \ref{sec_proposed_model}. The proposed inference algorithm is discussed in section \ref{sec:inference}. To demonstrate the effectiveness of the proposed model, extensive experimental results are reported and analyzed in section \ref{sec:experiments}. Finally, section \ref{sec:conclusion} concludes this paper and discusses paths for future research.

\section{Background}\label{sec:SettingsAndBackground}
We aim to infer the users' interests and their influence on each other by analyzing the contents being propagated over the network. To this end, we use dependent nonparametric models and point processes to jointly model the occurring time and topics of the events. For the sake of self-sufficiency, in this section, we review some necessary background on non-exchangeable nonparametric models and temporal point processes.

\subsection{Dependent Nonparametric models}
A Bayesian nonparametric model, is a Bayesian model with an infinite-dimensional parameter space. 
Dependent nonparametric models extend traditional models to define a probability measure over a set of dependent measures or clusterings usually indexed by a covariate \cite{DependentNPModelsSurvey}. 

For example, Recurrent Chinese Restaurant Franchise Process (RCRFP) is a dependent nonparametric model for clustering dependent groups of data \cite{ahmed2012timeline}.  This process assumes that data is categorized into a set of disjoint groups and the data in each group is exchangeable. However, it is assumed that the groups are indexed by a covariate such as time and are dependent of each other. In this model, the number of clusters is unknown and hence RCRFP infers the number of clusters in each group and simultaneously clusters them in to a set of shared clusters to capture the latent structure of each group. For example, in our problem RCRFP can be used to cluster the set of events of different users. Moreover, since the people are interested in a set of common topics, RCRFP shares the clusters among the users.  Although this model is a good match for clustering the events over a network,  the exchangeability of events of each user is not a valid assumption in this problem. In Section \ref{sec_proposed_model}, we propose an extended version of RCRF that also models the dependency among the customers of a restaurant.

\subsection{Temporal Point Processes}

Temporal point processes are a set of powerful methods for modeling a list of time-stamped events $(t_1, \ldots, t_n)$. 
A temporal point process can be completely specified by distribution of its inter-event times \cite{Daley2002}:
\begin{align}
f(t_1, \ldots, t_n) = \prod_{i=1}^n f(t_i | t_1, \ldots, t_{i-1}) = \prod_{i=1}^n f^*(t_i)  
\end{align}
To specify a point process, it suffices to define  $f^*(t) $, or equivalently $f(t|\mathcal{H}_t)$, where $\mathcal{H}_t$ is the history of events up to time $t$. A more intuitive way to characterize a temporal point process is to define the conditional intensity function \cite{Aalen2008}, which is defined as:
\begin{align} \label{eq:lambda}
\lambda^*(t) = \frac{f^*(t)}{1-F^*(t)}
\end{align}
where $F^*(t)$ is the CDF of $f^*(t)$. 
Different point processes can be determined by specifying appropriate intensity functions. For instance, in a homogeneous Poisson process, the intensity is independent of the history, and is constant over time, i.e. $\lambda^*(t) = \lambda$ \cite{Kingman1992}. 


In order to model the events of multiple dependent sources, multidimensional point processes can be utilized. In a multidimensional point process, the intensity of a dimension depends on the event history of all dimensions. 

Each event can also be associated with some auxiliary information. This information is known as the mark of an event, and the associated point process is called a marked point process. For example, the topics of tweets propagated through a network can be considered as the marks of events. Dependent nonparametric models are a set of flexible tools for modeling marks of events that can adapt their complexity according to data. These models can become a powerful tool for modeling temporal data when combined with point processes. Moreover, these models can become more flexible if the complexity of intensity function can be adapted to the complexity of temporal data. In the next section, we describe HNP3, which is a nonparametric multidimensional point process.

\section{Proposed Model}\label{sec_proposed_model}
In order to model the propagation of content over a social network, we propose the Hierarchical NonParametric Point Process (HNP3). HNP3 is a framework for modeling the event histories of a group of dependent sources, in which the topics are shared among the sources and the number of topics is unbounded. The main idea of HNP3 is to use a multidimensional point process to model the time of events and a hierarchical nonparametric model to model the marks of events. 

Let $\mathcal{D}(t) = \{e_i\}_{i=1}^{N(t)}$ denote the set of events observed until time $t$, where the event $e_i$ is a triple $(t_i,u_i,d_i)$ which indicates that at time $t_i$, user $u_i$ shares document $d_i$. 
Since the members of a network influence each other, the events in a network are mutually-exciting, i.e. each event triggers some new events in the network.  Hence, the events can be categorized into endogenous and exogenous events. Endogenous events are the responses of users to the actions of their neighbors within the network, and exogenous events are user actions based on external drivers. Let $s_i$ denote the triggering event for event $i$. If the event is exogenous, then $s_i=i$ and otherwise it is the index of event that has triggered the $i$th event.

Each event $e_i$ also has a corresponding latent topic $\theta_i$ which is regarded as its mark. We assume that the topic of an endogenous event is the same as the topic of its triggering event. Moreover, we assume that each user $u$ has a distribution $G_u^t$ over the topics at time $t$ that represents his interest over different topics, and he selects the topic of an exogenous event randomly from this distribution at time $t$. 

Since the users of a network are usually interested in a set of common topics, we assume that the favorite topics are shared among the users. Let  $\{\phi_k\}_{1}^{K(t)}$ denote the set of unique topics over the network until time $t$. Each user $u$ is interested in a subset of these topics at any time $t$ which is denoted by $\{\psi_{ui}\}_{i=1}^{K_u(t)}$, where $K_u(t)$ denotes the number of topics that user $u$ is interested in, at time $t$.
 
Each topic is a distribution over the words of the dictionary. Every document with topic $\theta$ has the same distribution over the words as the distribution of $\theta$. Moreover, we assume that each topic has a specific temporal dynamic which shows the rate at which the events of that topic diffuse over the network.

\begin{figure*}[t]
  \includegraphics[width=\textwidth]{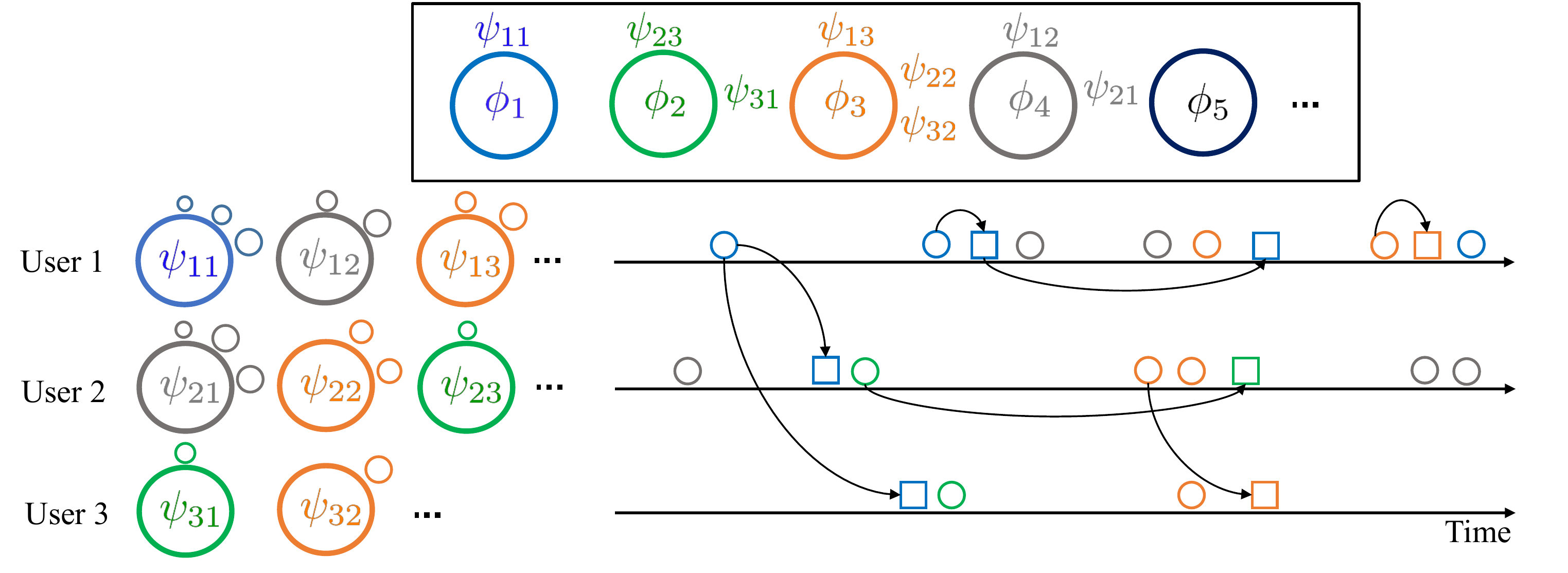}
  \caption{The illustration of HNP3 model: The top level restaurant represent the popularity of topics over the network. The interest of each user corresponds to distribution over the topics which is represented by a restaurant. The popularity of each topic is the weighted sum of the exogenous events generated by the user. Exogenous and endogenous events are represented by circles and squares, respectively. The arrows show the triggering relationship among events.}\label{fig:HierarchicalClustering}
\end{figure*}

As depicted in Fig. \ref{fig:HierarchicalClustering}, we propose a three-level nonparametric model for clustering events according to their topic. In the first level, the Hawkes process clusters the events based on their triggering event.
We use a variation of RCRFP to cluster the exogenous events of each user in the second level and share the topics among all users in the last level.

For clarity, we use the following notation for the remainder of this paper. $\mathcal{D}_{uk}^s(t)$ denotes the set of events triggered by event $s$ generated by user $u$ until time $t$ with topic $\phi_k$. Let $D^0(t)$ be the set of exogenous events until time $t$. We use dot notation to represent union over the dotted variable, \textit{e.g.}, $\mathcal{D}_{u\Cdot}(t)$ represent the events of user $u$ before time $t$ with any topic, and $\mathcal{D}_{\bar{u}k}(t)$ represent the events of all users except $u$, before time $t$, with topic $k$. Moreover, let $z_i$ be the index of the topic of the $i$th event among $\phi_k$s. That is, $\theta_i=\phi_{z_i}$. 

\subsection{The Proposed Generative Model}
We assume that the time at which user $u$ publishes documents follows a Hawkes process with 
intensity function:
\begin{align}\label{eqn:totalLambda}
	\lambda_{u}(t)= \mu_u + \sum_{s=1}^{N(t)-1}{\lambda_u(t,s)}
\end{align}
where $\mu_u$ is the exogenous intensity which shows the tendency of user $u$ to generate new events. $N(t)$ represents the number of events until time $t$. $\lambda_u(t,s)$ is the amount of intensity of user $u$ at time $t$ that is caused by event $s$.  $\lambda_u(t,s)$ is defined as $\alpha_{u_su}\kappa_{z_s}(t,t_s)$ where $\alpha_{u_su}$ is the influence of event $s$'s user on $u$. $\kappa_{z_s}(t,t_s)$ is a kernel function which determines the diffusion rate of events with topic $z_s$. In our case, we use the exponential kernel:
\begin{align}
	\kappa_{k}(t,t_s)=e^{-\beta_{k}(t-t_s)}
\end{align}

As it was mentioned before, we use the topic of the document as the mark of events. Using the aforementioned assumptions, if the event $e_i$ is exogenous and we know the triggering event $s_i$, then the topic of $e_i$ is the same as the topic of $e_{s_i}$, i.e. $\theta_i=\theta_{s_i}$. Otherwise, the user $u_i$ selects one of his previously used topics $\psi_{uj}$ with probability $\frac{n_{uj}(t)}{n_{u:}(t)+\gamma}$ or selects a new topic with probability $\frac{\gamma}{n_{u:}(t)+\gamma}$:
\begin{align}
	p(\theta_{i}|&u_i=u,s_i=i,t_i,z_{1:i-1})= \\\nonumber
	&\sum_{k=1}^{K_u(t)} \frac{n_{uk}(t)}{n_{u\Cdot}(t)+\gamma} \delta(\psi_{uk})+\frac{\gamma}{\gamma+n_{u\Cdot}(t)}\delta(\psi_{u,new}^t)
\end{align}
where $\gamma$ is a parameter which shows the tendency of users to talk about new topics, and $n_{uk}(t)$ is the weighted number of exogenous events of user $u$ with topic $\psi_{uk}$, that is:
\begin{align}
	n_{uk}(t)=\sum_{e\in D_u^0(t)}\exp(-\nu(t-t_e))\mathcal{I}(\theta_{e}=\psi_{uk})
\end{align}
where $\exp(-\nu(t-t_e))$ is a kernel which represents the decaying impact of events over time. In order to share the topics among the users, we use the same idea as RCRF and assume that users select their new topics from a common discrete distribution which shows the popularity of topics over the network:
\begin{align}\label{eqn:G_0}
		p(\psi_{u,new}^t&|\psi_{\Cdot\Cdot} ,\gamma,H)=\\\nonumber
		&\sum_{l=1}^{K(t)} \frac{m_{k}(t)}{\zeta+m_{\Cdot}(t)} \delta(\phi_l)+\frac{\zeta}{\zeta +m_{\Cdot}(t)}H
\end{align}
where $m_k(t)$ shows the popularity of topic $\phi_k$ over the whole network, and is the weighted number of times users select a new topic $\phi_k$ from \ref{eqn:G_0}, that is:
\begin{align}
	m_k(t) =  \sum_{e\in D^0(t)}\exp(-\nu(t-t_e))\mathcal{I}(\theta_{e}=\phi_{k},l_e=1)
\end{align}
where $l_e=1$ indicates that the topic of the exogenous event $e$ is a new one and is sampled from \ref{eqn:G_0}.

Finally, we draw the content of a document from the distribution of its topic over the words of dictionary:
$$d_{i}|\phi_{1:K(t)},z_{i} \sim Mult(\phi_{z_i})$$
\section{Inference}\label{sec:inference}
%
We use a two-step iterative algorithm to update our beliefs about the latent variables in an online manner. First, we use collapsed Sequential Monte Carlo (SMC) \cite{smith2013sequential} to estimate the posterior distribution of local latent variables by marginalizing out all global latent variables except $\beta_k$s. In the second step, we estimate $\beta_k$ using the learned distributions.
%
Each particle represent a hypothesis about the set of latent variables and its weight shows our confidence about it. By observing every new event, each particles is updated by appending a new $(s_{n+1},z_{n+1})$ to it, and updating their weights correspondingly. To this end, we need a proposal distribution $q(s_{n+1},z_{n+1}|s_{1:n},z_{1:n},\mathcal{H}_t)$ to sample from. In order to minimize the variance of the weights, we use its posterior, \cite{ahmed2011online} i.e. $p(s_{n+1},z_{n+1}|s_{1:n},z_{1:n},\mathcal{H}_t)$. 
We assume a Gamma prior over the betas. In order to compute the expected value of its posterior, we draw $M$ samples from the prior and find the mean as follows:
\begin{align}
	\mathbf{E}\left[\beta_k|t_{1:N},z_{1:N},s_{1:N}\right]\approx \sum_{m=1}^Mw_m\beta^{(m)}
\end{align}
where $w_m$ is the weight of $m$th sample and is proportional to likelihood $p(t_{1:N},z_{1:N},s_{1:N}|\beta_m)$.
In the next section we show the effectiveness of the proposed inference algorithm by several experiments on synthetic and real data.

\section{Experimental Results}\label{sec:experiments}
In this section, we empirically evaluate the performance of HNP3 by using both synthetic and real data. 
The experiments on synthetic data are used to evaluate the effectiveness of the inference algorithm introduced in section \ref{sec:inference}. For the real data, we investigate the performance of HNP3 model in inferring the hot topics over the network and their corresponding temporal dynamics. Moreover, we evaluate its power to predict the time of next events and also inferring the influence network.

\begin{figure*}[!t]
\centering
\includegraphics[width=\textwidth]{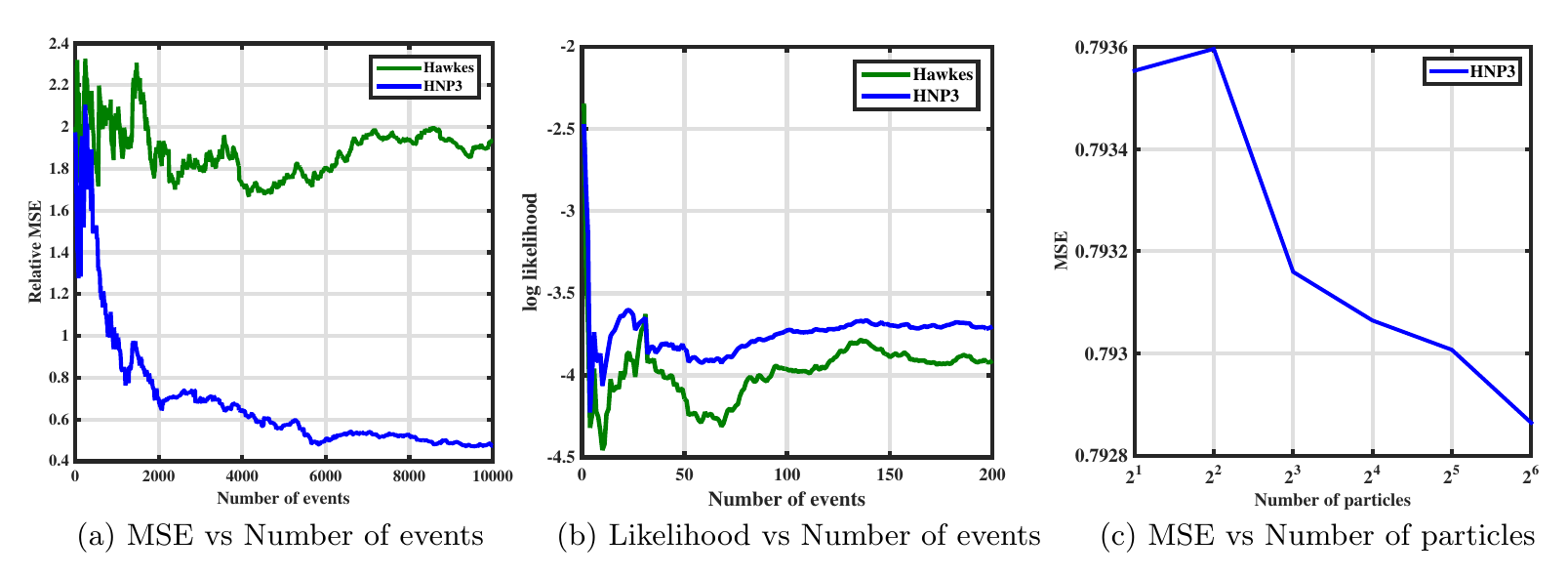}
\caption{The performance of HNP3 method on synthetic data. Figure (a) shows the relative error in estimating the influence matrix and exogenous intensity parameter. Part (b) compares the mean log likelihood of time for next events in HNP3 and Hawkes models. Figure (c) shows the error in estimating the influence matrix with different number of particles.}
\label{fig:synthetic}
\end{figure*}
\subsection{Synthetic Data}
In order to evaluate the performance of the proposed inference algorithm, we generated a set of $10^4$ events by using the proposed generative model. We used the exponential kernel for all four topics with different $\beta$ parameters. Figure \ref{fig:synthetic}(a) shows the performance of HNP3 in estimating the influence matrix $\alpha$, and exogenous intensity parameters $\mu_u$s. 
As it is evident in Figure \ref{fig:synthetic}(a), although in the first $1000$ events, HNP3 does not make a significant improvement over the Hawkes method, but after learning the topics and their corresponding kernel, the error considerably decreases. 

Since the ability of HNP3 in predicting the time of future events heavily depends on correctly estimating the topics kernel, we compared HNP3 and Hawkes process based on the mean likelihood of time of next events, to confirm the efficiency of the proposed algorithm in learning the kernels. As it is depicted in Figure \ref{fig:synthetic}(b), the likelihood of the time of future events is consistently more than the Hawkes process. 

In order to determine number of particles in the inference algorithm, we tested the algorithm with different number of particles. As it is depicted in Figure \ref{fig:synthetic}(c), the precision of the algorithm in estimating the parameters of the model does not depend on the number of particles too much. Therefore, we used $8$ particles in all of our experiments.

\subsection{Real Data} 
We also evaluated performance of the proposed method on a real dataset, gathered from EventRegistry \footnote{\url{http://eventregistry.org/}}. 
For the real data, we first analyze the performance of HNP3 on modeling the content of events. To this end, we try to address the following questions: 1) How well HNP3 can capture different topics?, and 2)
How well HNP3 can capture the temporal dynamics of topics?
We also analyze the performance of HNP3 on predicting the time of next events and compare its performance with two well known state of the art methods.
\subsubsection{Dataset Description}
Our real dataset corresponds to articles extracted from EventRegistry, which is an online aggregator of news articles around the world. We have collected news articles containing each of 3 different tags; \textit{FIFA}, \textit{Iran-Sanctions}, and \textit{Paris-Attack} from 2015/11/01 to 2016/01/13. The collected data contains about $100000$ news articles and $100$ different news sites. The sites are treated as nodes and the articles as events. We have preprocessed the data and removed some stop-words and irrelevant words and extracted the bag of words for each article. 
\subsubsection{Results}
\begin{figure}[!t]
\centering
\subfloat[]{\includegraphics[width=0.75in]{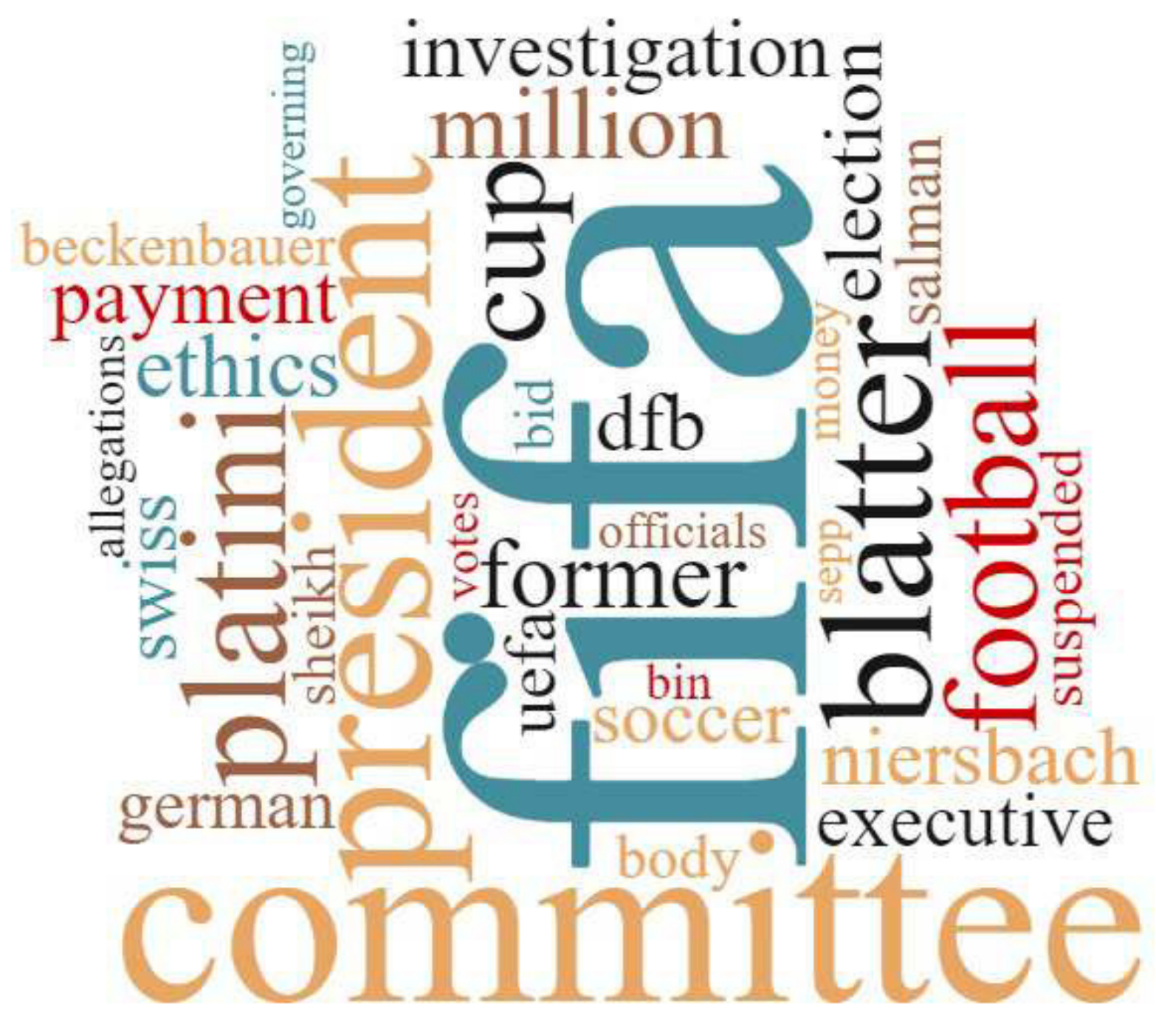}%
\label{fig:fifa_cloud}}
\hspace{0.2in}%
\subfloat[]{\includegraphics[width=2in]{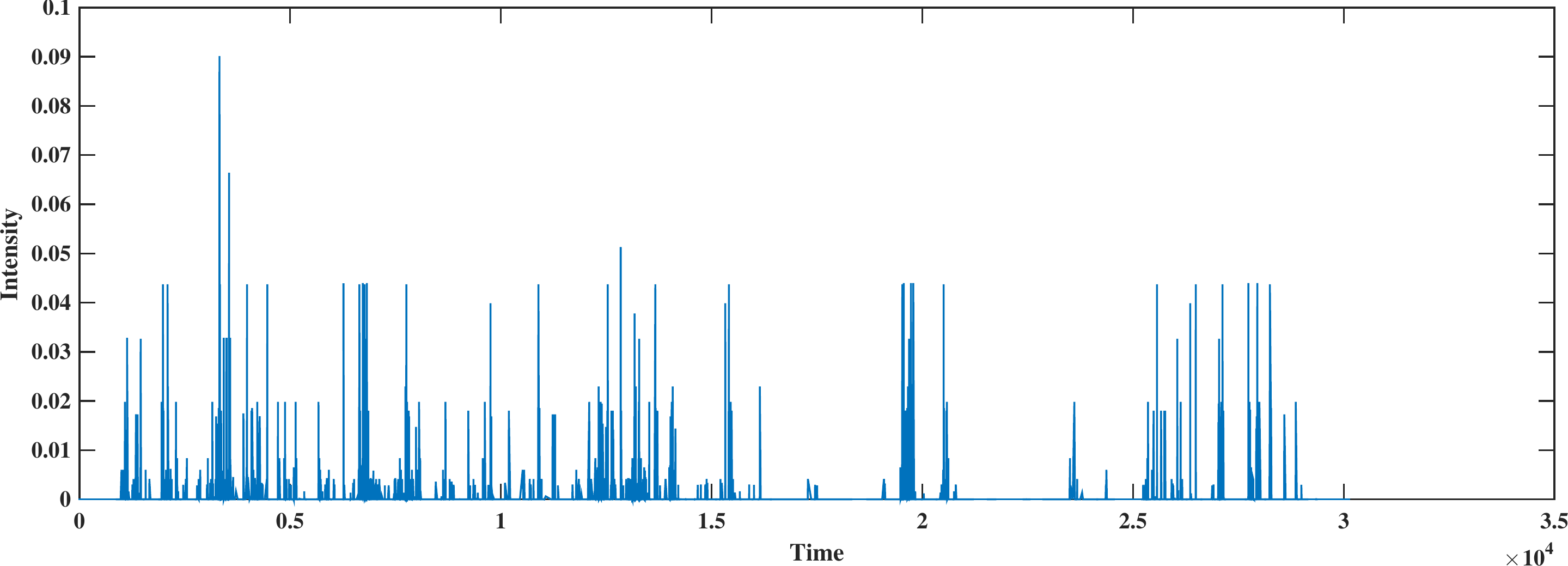}%
\label{fig:fifa_intensity}}
\vfil
\subfloat[]{\includegraphics[width=0.75in]{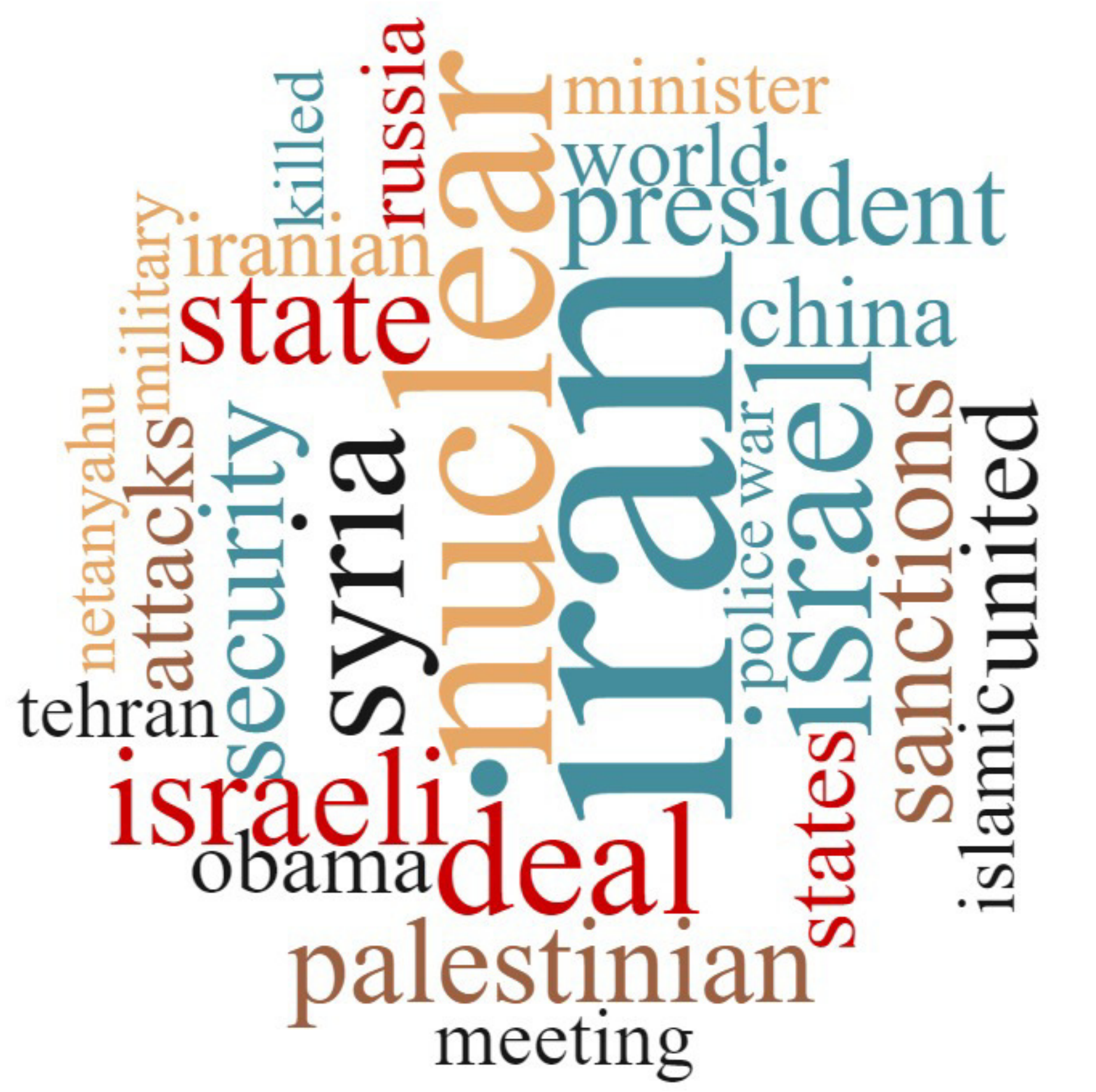}%
\label{fig:iran_cloud}}
\hspace{0.2in}%
\subfloat[]{\includegraphics[width=2in]{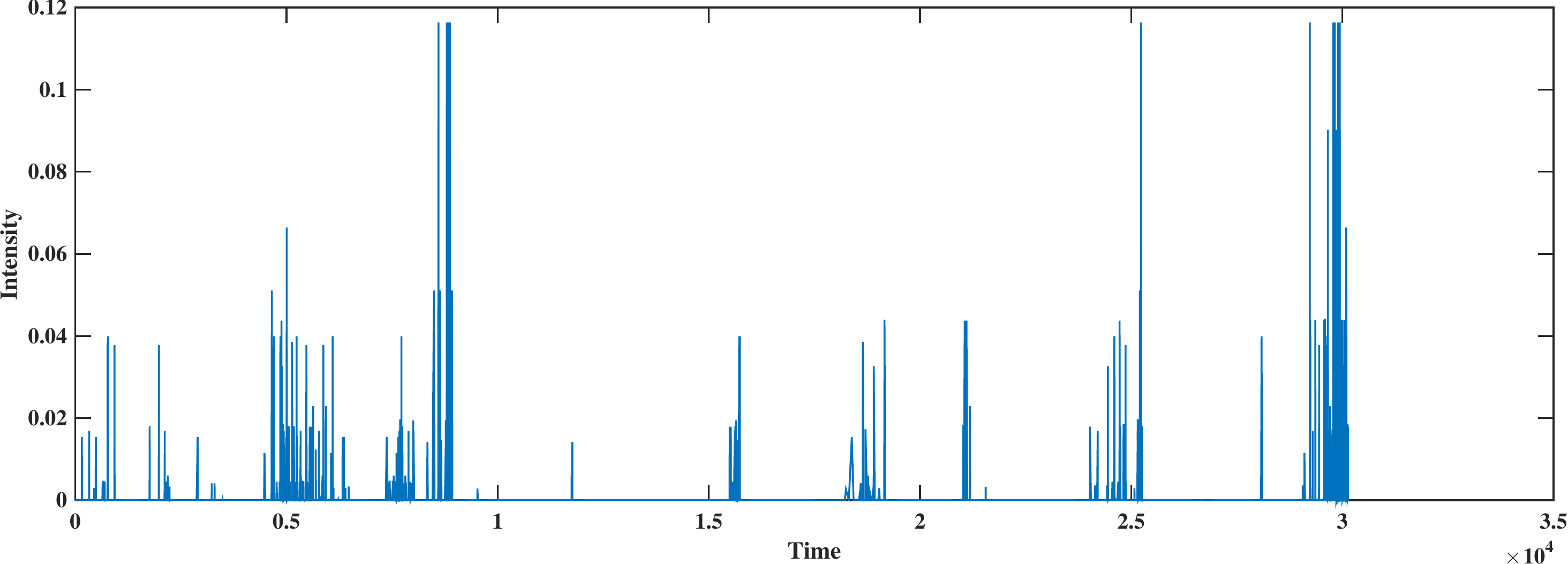}%
\label{fig:iran_intensity}}
\vfil
\subfloat[]{\includegraphics[width=0.75in]{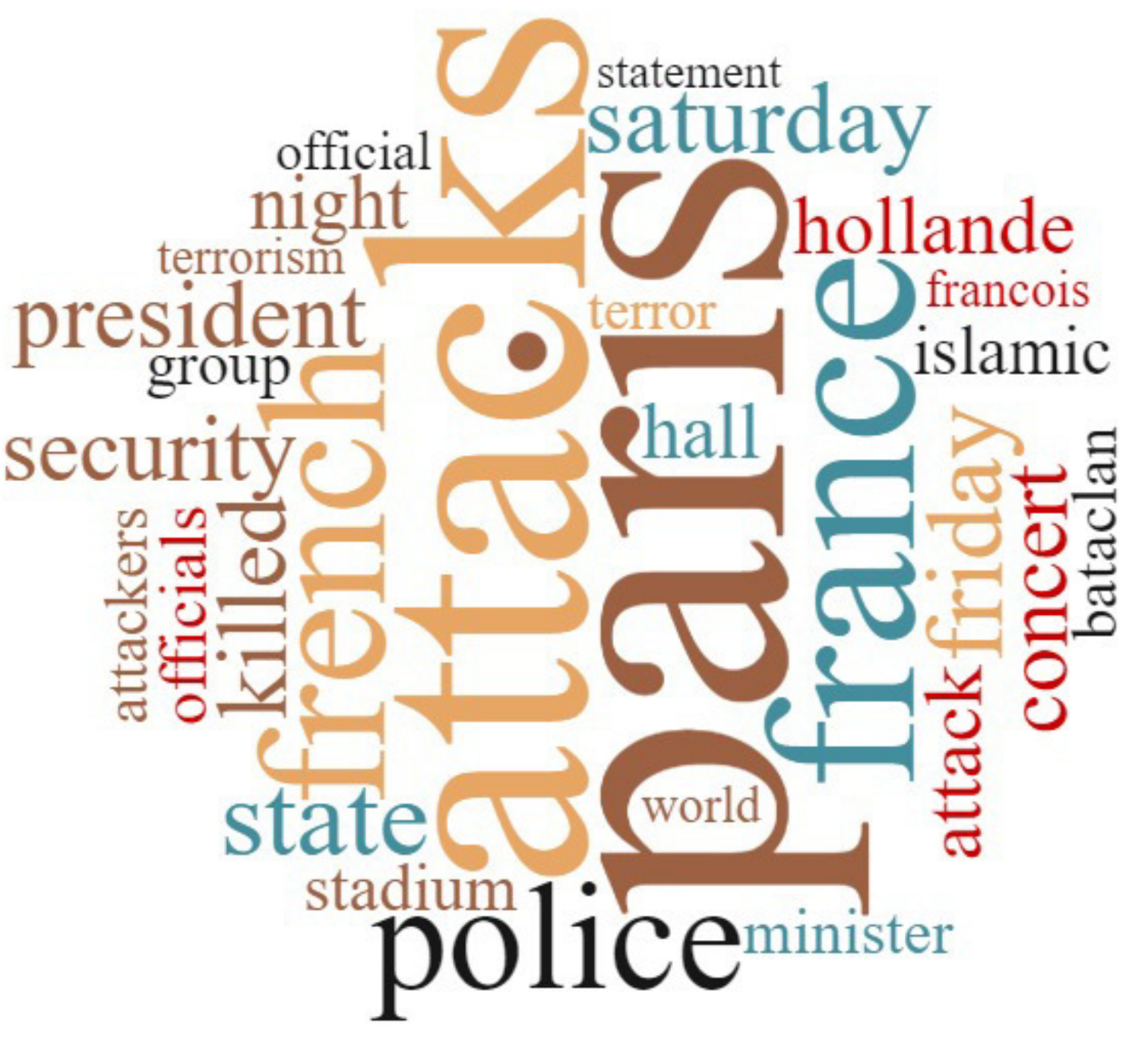}%
\label{fig:paris_cloud}}
\hspace{0.2in}%
\subfloat[]{\includegraphics[width=2in]{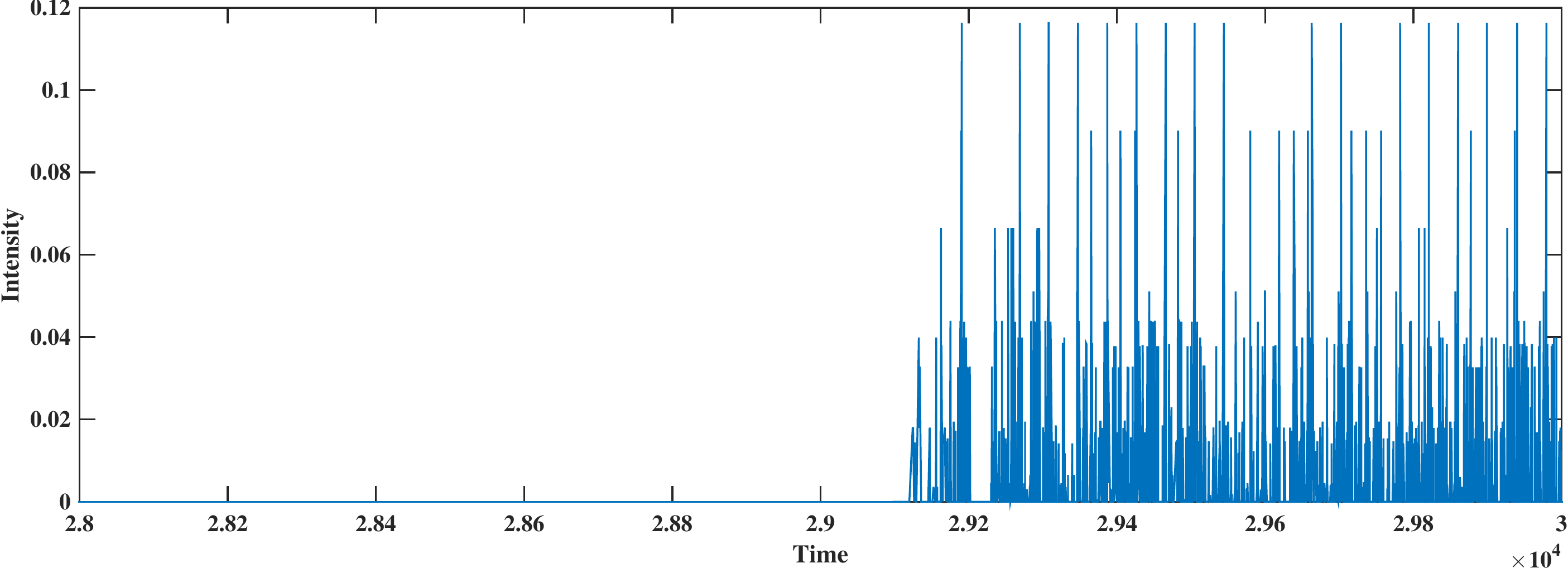}%
\label{fig:paris_intensity}}
\caption{Three main topics extracted by HNP3 from the EventRegistry dataset. For each topic, we show the word cloud of top frequent words in the first column. The second column represents the intensity function of each extracted topic, capturing its popularity dynamics, against time.}
\label{fig:real_content}
\end{figure}

\textbf{\\Content Analysis.}
To show the performance of HNP3 on detecting different topics, we depicted the top frequent words in 3 main topics discovered by HNP3. Figures \ref{fig:fifa_cloud}, \ref{fig:iran_cloud}, and \ref{fig:paris_cloud} shows the word cloud of top frequent words in 3 main topics learned by HNP3. As it can be seen, HNP3 can detect meaningful clusters which are representative of true real topics and represent corresponding events.

To analyze the temporal dynamics of different topics, we depicted the intensity function of each topic against time, which is representative of their popularity over time. Figures \ref{fig:fifa_intensity}, \ref{fig:iran_intensity}, and \ref{fig:paris_intensity} represents the intensity function of 3 different detected topics over time. The results show some interesting patterns that confirm the good performance of HNP3 in capturing temporal dynamics of popularity for different topics. As it can be seen from Fig. \ref{fig:paris_intensity}, the popularity of \textit{Paris-Attack} topic rises suddenly somewhere in time. This is reasonable, since we collected data two weeks before the \textit{Paris-Attack} event. Therefore, the intensity of events is zero before the event, and suddenly rises after a large number of events are generated after it happens. Since the \textit{FIFA} topic is discussed all the time, its intensity is also evenly distributed over the time axis. The \textit{Iran-Sanctions} topic also has a periodical popularity pattern. Since the negotiations about Iran sanctions took place periodically, it is desirable that its popularity rises just after these negotiations and then fades out. 

The above results indicate that the HNP3 performance is acceptable on detecting different topics, capturing their triggering kernels, and their temporal dynamics over time.
 
\begin{figure}[!t]
\centering
\includegraphics[width=0.45\textwidth]{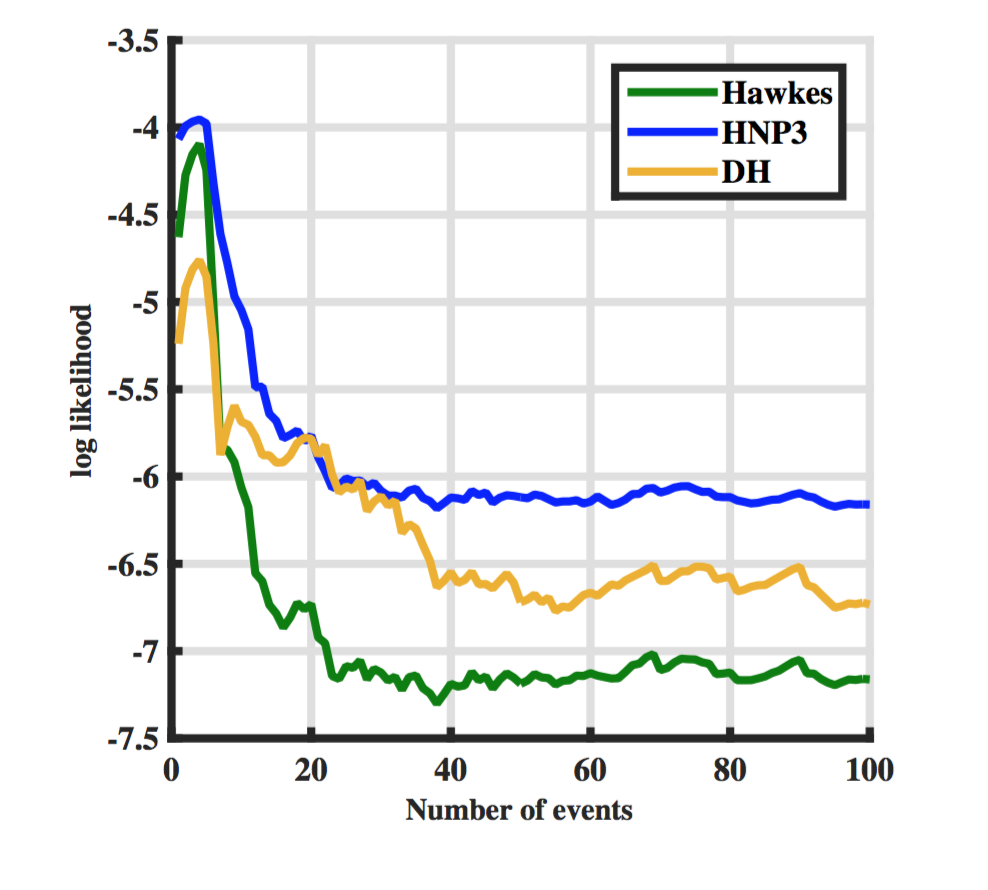}
\caption{The performance of different methods on predicting time of next events for the EventRegistry data.}
\label{fig:real_time_likelihood}
\end{figure}

\textbf{Prediction.} 
We also compared the performance of HNP3 on predicting the time of next events with the Hawkes and Dirichlet-Hawkes(DH) models. To this end, we trained each model with some events, and computed the time likelihood of next events for each model. Fig. \ref{fig:real_time_likelihood} represents the likelihood of next 100 events for HNP3, Hawkes, and DH models. 
As it is shown in Fig. \ref{fig:real_time_likelihood}, HNP3 performs better than the Hawkes and DH models. Moreover, it can be seen that the HNP3 and DH models which utilize the content of events, perform better than the Hawkes model which ignores the content. 
We also observe that the HNP3 model which considers the network effect and the influence of friends, performs better than the DH model which do not consider the influence of users on each other.

\section{Conclusion}\label{sec:conclusion}
In this paper, we introduced a framework for modeling dependent groups of temporal events with complex longitudinal dependencies. This framework is able to jointly model the time and marks of events and adapt itself to the complexity of data. The framework also provides a  hierarchical clustering of the events by utilizing the dependency among content and time of events. This clustering may have many applications in different areas. For instance, we used the framework for modeling the content diffusion over social media and the clustering allowed us to infer the source of events and also the hot topics over the network. 

Moreover, HNP3 uses multidimensional point processes for modeling time of events. However, the intensity function of this process is a mixture of intensities and its complexity grows with the number of data. In addition, HNP3 utilizes dependent nonparametric methods for modeling marks of events. These capabilities allow HNP3 to adapt its temporal and topical complexity according to the complexity of data, which makes it a suitable candidate for real world scenarios. 



Since diffusion of contents over networks has gained a lot of attention in recent years, we applied HNP3 to this real application and designed an online inference algorithm based on SMC, which can efficiently infer parameters of the model.
Experiments on synthetic data showed the efficiency of our inference algorithm. The experimental results on real data confirmed the superior performance of the proposed method compared to other recent methods in finding different topics and their diffusion rates.

There are many lines to extend this study. For example, we used Hawkes process for modeling time of events. One plan to extend this method is to use more complex point processes that are analogous to more complex clustering algorithms such as hierarchical dd-CRP \cite{ghosh2014nonparametric}. 


\newpage
{	
\bibliographystyle{IEEEtran}
\bibliography{references}
}

\end{document}